\documentclass{article}

\usepackage[preprint]{neurips_2024}

\usepackage[utf8]{inputenc} % allow utf-8 input
\usepackage[T1]{fontenc}    % use 8-bit T1 fonts
\usepackage{hyperref}       % hyperlinks
\usepackage{url}            % simple URL typesetting
\usepackage{booktabs}       % professional-quality tables
\usepackage{amsfonts}       % blackboard math symbols
\usepackage{nicefrac}       % compact symbols for 1/2, etc.
\usepackage{microtype}      % microtypography
\usepackage{xcolor}         % colors
\usepackage{graphicx}
\usepackage{amsmath}

\usepackage{float}

\title{1st-Order Magic: Analysis of Sharpness-Aware Minimization}

\author{
  Nalin Tiwary\thanks{Equal contribution} \quad 
  Siddarth Aananth\footnotemark[1] \\
  University of Illinois at Urbana-Champaign \\
  {\texttt{\{nalint2,aananth2\}@illinois.edu}} \\
}

\begin{document}

\maketitle

\begin{abstract}
Sharpness-Aware Minimization (SAM) is an optimization technique designed to improve generalization by favoring flatter loss minima. To achieve this, SAM optimizes a modified objective that penalizes sharpness, using computationally efficient approximations. Interestingly, we find that more precise approximations of the proposed SAM objective degrade generalization performance, suggesting that the generalization benefits of SAM are rooted in these approximations rather than in the original intended mechanism. This highlights a gap in our understanding of SAM's effectiveness and calls for further investigation into the role of approximations in optimization.
\end{abstract}

\section{Introduction}

Modern machine learning relies heavily on overparameterization to achieve state-of-the-art performance across various tasks. However, with this overparameterization comes the risk of overfitting, where models can easily memorize training data, even when the labels are random \cite{overifitting}. To mitigate this risk, effective training techniques must minimize training error in a way that also ensures good generalization.

One promising direction for improving generalization is focusing on the geometry of the loss landscape, particularly sharpness. Previous studies have demonstrated that flatter minima in the loss landscape are often associated with better generalization. \cite{keskar_flat_minima_generalization} showed that larger batch sizes can lead to sharper minima and poorer generalization. This insight, along with the work of others such as \cite{dinh_sharpness_generalization} and  \cite{neyshabur_sharpness_generalization}, has motivated the development of optimization techniques that explicitly address sharpness.

Building on these ideas, several Sharpness-Aware optimization methods have been introduced. For instance, \cite{sam}, \cite{asam}, and \cite{zheng2021regularizing} leverage the connection between sharpness and generalization by minimizing the loss at perturbed parameter points, thereby penalizing sharp regions of the loss landscape. Despite the empirical success of these methods, the underlying mechanisms behind SAM's effectiveness remain unclear.

Specifically, the Sharpness-Aware Minimization (SAM) objective aims to penalize a specific notion of sharpness, but the practical implementation of SAM involves a series of approximations. These approximations, including the use of a first-order Taylor expansion, imply assumptions about the local linearity of the loss landscape around the current parameters, which are often inaccurate. Surprisingly, improving these approximations does not lead to better generalization \cite{sam_ga}. 

% \siddarth{We need some sort of citation for this, to show that this is true}

In this paper, we analyze SAM's performance when using better approximations of its proposed objective, explore the reasons behind the observed discrepancy in generalization, and provide a standardized sharpness comparison. We also investigate the approximations involved in the SAM objective and, specifically, the role the 1st-order Taylor expansion approximation plays in the superior generalization of the 1st-order formulation of SAM. We propose a modification, Rand-SAM, which empirically demonstrates our proposed reasoning behind the seemingly "magical" performance benefits of the original SAM formulation.

\section{Related Works}
% - Flatter minima generalize better
% - SAM itself introduce objective formation
% - GA ascent paper
% - Tengyu Ma paper

\subsection{Sharpness-Aware Minimization} \label{sec:sam}

\begin{figure*}[ht]
    \centering
    \includegraphics[width=0.8\linewidth]{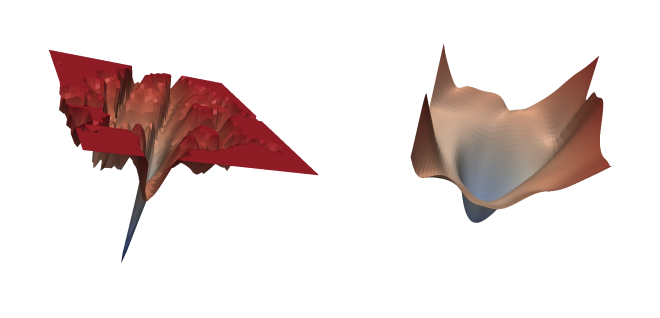}
    \caption{The figure to the left is a sharp minima obtained through SGD while the figure to the right is a flat minima obtained through SAM}
    \label{fig:sharp vs flat minima}
\end{figure*}

We focus on the Sharpness-Aware Minimization objective introduced in \cite{sam}. 

\begin{equation}
    \mathcal{L}^{SAM}(w) = max_{\|\epsilon\|\leq \rho}\mathcal{L}(w+\epsilon)
\label{eq:sam_obj}    
\end{equation}

In this formulation, $w \in \mathbb{R}^{d}$ represents the model parameters, and $\mathcal{L}$ is the training loss. The parameter $\rho$ controls the size of the neighborhood used to penalize sharpness, typically set to $0.05$. The SAM objective aims to find parameter values where the entire neighborhood of radius $\rho$ exhibits low training loss, leading to flatter minima. Figure. \ref{fig:sharp vs flat minima} illustrates an example of such a flat minimum.

In order to make the computation efficient, \cite{sam} propose a series of approximations. The approximations of interest to us occur while obtaining the value of $\epsilon$

\begin{equation}
    \epsilon^*(w) = argmax_{\|\epsilon\|\leq \rho}\mathcal{L}(w+\epsilon)
\label{eq:ideal_eps}    
\end{equation}

Since directly solving Eq. \ref{eq:ideal_eps} is computationally expensive, SAM approximates $\epsilon^*$ using a first-order Taylor expansion around $w$ w.r.t $\epsilon$ around 0:

\begin{equation}
    \epsilon^*(w) \approx \epsilon_1(w) = argmax_{\|\epsilon\|\leq \rho}(\mathcal{L}(w)+\epsilon^T \nabla_w \mathcal{L}(w))=argmax_{\|\epsilon\|\leq \rho}\epsilon^T \nabla_w \mathcal{L}(w)
\label{eq:first_order_taylor_eps}    
\end{equation}

Since Eq. \ref{eq:first_order_taylor_eps} is maximizing a linear function under a norm\footnote{We utilize L2 norm unless specified otherwise and gradient w.r.t to $w$ unless specified otherwise} constraint, $\epsilon_1$ will be a vector with norm $\rho$ is the same direction as $\nabla L(w)$ as shown in Eq. \ref{eq:first_order_eps}

\begin{equation}
    \epsilon_1 = \rho \frac{\nabla L(w)}{\|\nabla L(w)\|}
\label{eq:first_order_eps}    
\end{equation}

Utilizing $\epsilon$, \cite{sam} approximates $\nabla L^{SAM}(w)$ as $\nabla L(w+\epsilon_1)$ and utilizes gradient descent.

\subsection{Gradient-Ascent}

\cite{sam} proposes $N$-step Gradient Ascent as an alternate way to estimate $\epsilon^*$. For each iteration $n \in \{1,2,..,N\}$ and $w^N=w+\epsilon_N$

\begin{equation}
    w^n=w^{n-1} + \frac{\rho}{N} \frac{\nabla L(w^{n-1})}{\|\nabla L(w^{n-1})\|}
\label{eq:ga}
\end{equation}

\cite{sam_ga} discusses this formulation of SAM and the impact of the value of $N$ on the loss landscape and specifically addresses the phenomenon that an increase in N, which should provide a better approximation of $\epsilon^*$, rarely improves generalization performance.

\subsection{Notion of Sharpness} \label{sec:notions of sharpness}

\begin{table}[ht]
\centering
\caption{Notions of sharpness}
\begin{tabular}{c c c}
\hline
\textbf{Type of Loss} & \textbf{Notation} & \textbf{Definition}   \\ \hline
Worst-direction & $L^{\text{Max}}$ & $\max_{\|\epsilon\| \leq \rho} L(w + \epsilon)$ \\ 
Ascent-direction & $L^{\text{Asc}}$ & $L\left(w + \rho \frac{\nabla L(w)}{\|\nabla L(w)\|}\right)$\\
Average-direction & $L^{\text{Avg}}$ & $\mathbb{E}_{g \sim N(0, I)} L\left(w + \rho \frac{g}{\|g\|_2}\right)$ \\ \hline
\end{tabular}

\label{tab:notions_of_sharpness}
\end{table}

As pointed out in \cite{sam_sharpness_notion}, the proposed SAM objective ($L^{\text{Max}}$) and the actual SAM objective ($L^{\text{Asc}}$) use subtly different notions of sharpness due to the approximations mentioned in Sec. \ref{sec:sam}. Both of these notions of sharpness differ from the notion of sharpness used to upper bound the generalization error of SAM ($L^{\text{Avg}}$), as shown in Table. \ref{tab:notions_of_sharpness}.

\cite{sam_sharpness_notion} also utilizes a gradient flow-based analysis, in the style of \cite{gradient_flow}, to analyze the biases introduced by these notions of sharpness and utilizes this analysis to provide theoretical underpinnings for the mechanism behind the different formulations of SAM from \cite{sam}. Since the analysis is based on gradient flow, it assumes $\rho \rightarrow 0$, which is not appropriate for SAM since $\rho = 0.05$. Furthermore, with the first-order assumption being made for each discrete step in the algorithm, the continuous nature of gradient-flow analysis introduces further differences. So, while this analysis does shed light on the role of the approximations shown in Sec. \ref{sec:sam} in the generalization success of SAM, it does not paint the complete picture.

\section{Intuition and Analysis}
% \siddarth{We need another name for this: Methodology and Hypothesis maybe}
% \siddarth{If Neurips does not have a fixed set of names, lets just call this Hypothesis, idts that this section has anything to do with a method since there is no specific method we are proposing, just claims and imperial evidence for the claims}
\label{sec: intuition}

To demonstrate that the approximations made in the SAM objective, leading to Eq. \ref{eq:first_order_eps}, are the driving factor behind SAM's generalization success, we analyze SAM using $N$-step gradient ascent. Specifically, previous works \cite{sam, sam_ga} show that increasing $N$ improves the accuracy of the approximation of $\epsilon^*$. However, counter intuitively, this does not improve generalization. In fact, higher values of $N$ often result in significantly sharper solutions, with sharpness measured using Eq. \ref{eq:inaccurate_sharpness}.

\begin{equation}
    sharpness = \mathcal{L}(w+\epsilon_N)- \mathcal{L}(w)
\label{eq:inaccurate_sharpness}    
\end{equation}

One issue with comparing sharpness using this metric is that the accuracy of the $\epsilon^*$ approximation varies for different values of $N$. Therefore, a standardized measure of sharpness is needed to make fair comparisons between implementations. Nevertheless, the critical question remains: why does improving the approximation of $\epsilon^*$—leading to more accurate sharpness penalization—result in worse generalization?

To answer this question, we must look at Eq. $\ref{eq:first_order_eps}$. Specifically due to the reasons mentioned in Sec. \ref{sec:sam}, the condition $\|\epsilon_1\|= \|(w+\epsilon_1)-w\|=\rho$ implies that $\epsilon$ always lies on the boundary of the $\rho$-radius ball around the current parameter values $w$. Without additional assumptions about the loss landscape, there is no direct relationship between the loss at this boundary point ($w + \epsilon_1$) and the loss at the true optimal perturbation ($w + \epsilon^*$). Due to the complexity of the loss landscape, analyzing this relationship without overly restrictive assumptions becomes challenging.

We hypothesize that the success of the first-order SAM formulation stems from the fact that $\epsilon_1$ is always on the boundary of the neighborhood. By focusing on a boundary point, SAM penalizes high loss only at the boundary, which is characteristic of sharp minima, as shown in Figure. \ref{fig:sharp vs flat minima}. This selective penalization makes the first-order SAM formulation more robust to isolated spikes in training loss or training noise, effectively ignoring outliers within the $\rho$-radius neighborhood. Consequently, this leads to improved generalization, as SAM converges to flatter minima less influenced by local noise. 
% \siddarth{Mention something about better generalization as well as performance....maybe in experiments we can show generalization gap as well.}

\section{Experiments}

% \siddarth{Worry about the table positions and placement in the end - h, ht, H etc - after all sections are done}\\
% \siddarth{@Nalin: Recheck the data entry using the collab link and github logs}\\
% \siddarth{Need to cite datasets}
\subsection{N-step Gradient Ascent}
\label{sec: n-step}

As mentioned in Sec. \ref{sec: intuition}, increasing $N$ in $N$-step gradient ascent does not improve generalization and leads to significantly sharper solutions. The issue, however, is that methods, such as the calculation in the Appendix of \cite{sam}, compare sharpness using the same process without regard for the different values of $N$. The issue with that is that sharpness given by $\mathcal{L}(w+\epsilon_N)-\mathcal{L}(w)$ depends on $\epsilon_N$ which is dependent on the choice of $N$ as shown in Eq.\ref{eq:ga}.

We hence use a standardized notion of sharpness where we use the gradient ascent implementation in SAM but independently calculate sharpness utilizing the $\mathcal{L}^{Asc}$ notion of sharpness shown in Table. \ref{tab:notions_of_sharpness} utilizing Eq. \ref{eq:first_order_eps} since it has no dependence on $N$.

We hence re-create the experiments using this modified calculation of sharpness so that we can compare values for any arbitrary value of $N$. The results we get from Table. \ref{tab:wide_resnet_cifar_ga_c10}, substantiate our hypothesis that there is something 'magical' about SAM's first-order implementation. We observe that sharpness values are comparatively lower for $N = 1$ as compared to $N = 2, 3, 5$. 

Below in Table. \ref{tab:wide_resnet_cifar_ga_c10} we use a WideResNet-28-10 trained on CIFAR-10 \cite{alex2009learning} for 200 epochs, with implementations of SAM utilizing Gradient Ascent with $\rho = 0.05$ as shown in Eq. \ref{eq:ga} with $N=1,2,3,5$. Here, Sharpness is calculated as the difference between $\mathcal{L}(w+\epsilon_1)-\mathcal{L}(w)$ where $\epsilon_1$ is calculated using Eq. \ref{eq:first_order_eps} scaled by a factor of $10^3$ to ease readability. The generalization gap is the difference between training and testing loss, which we also scaled by $10^3$ to ease readability.

% \siddarth{Need to run 3-step SAM again as we need one more data point - only have 4 in analysis after removing the weird outlier}\\
% \siddarth{Need to maybe run 5-step SAM again as we need one more data point - only have 5 in total - but that is because I have no dropped any values}\\
% \nalin{Use $\mathcal{L}^{Asc}$ notation from Table. 
% \ref{tab:notions_of_sharpness}}\\
% \nalin{Talk about sharpness being significantly lower for 1st order SAM and maybe we can discard the generalization gap values}\\
% \siddarth{@Nalin run it on google colab}\\
% \siddarth{Either that or we need to use 3-4 values of average across the board...but I think 5 is ideal}\\

\begin{table}[H]
    \centering
    \caption{N-step Gradient Ascent formulated SAM - CIFAR10}
    % \siddarth{Ref above comments and google colab regarding n=3, n=5 issues}\\
    \label{tab:wide_resnet_cifar_ga_c10}
    \begin{tabular}{cccc}
    \hline
        \textbf{GA-Steps} & \textbf{Accuracy} & \textbf{Sharpness $\times 10^3$} & \textbf{Generalization Gap $\times 10^3$}    \\
    \hline
        1 &  $97.3_{\pm<0.1}$\% &  $4.8_{\pm2.7}$ & $47.3_{\pm1.1}$\\
        2 &  $97.3_{\pm<0.1}$\% &  $9.6_{\pm3.4}$ & $46.5_{\pm<1}$\\
        3 &  $97.3_{\pm<0.1}$\% &  $8.1_{\pm3.0}$ & $47.7_{\pm1.4}$\\
        5 &  $97.4_{\pm<0.1}$\% &  $9.2_{\pm3.4}$ & $46.4_{\pm1.2}$\\
        % 2 &  97.3\% &  9.7 & 52.6\\
        % 3 &  97.3\% &  6.0 & 47.6\\
        % 5 &  97.4\% & 9.2 & 46.4 \\
    \hline
    \end{tabular}
\end{table}

% \begin{table}[H]
%     \centering
%     \caption{N-step Gradient Ascent formulated SAM - CIFAR100}
%     \siddarth{ENTER VALUES}\\
%     \label{tab:wide_resnet_cifar_ga_c100}
%     \begin{tabular}{|cccc|}
%     \hline
%         \textbf{GA-Steps} & \textbf{Accuracy} & \textbf{Sharpness $\times 10^3$} & \textbf{Generalization Gap $\times 10^3$}    \\
%     \hline
%         1 &  TBD\% &  TBD & TBD\\ % 0.0444
%         2 &  TBD\% &  TBD & TBD\\
%         3 &  TBD\% &  TBD & TBD\\
%         5 &  TBD\% & TBD & TBD\\
%     \hline
%     \end{tabular}
% \end{table}

\subsection{Rand-SAM}

% \siddarth{Perhaps we need to cite SAM for SGD results in Rand-SAM experiments}\\
% \siddarth{@Nalin - lets re-run CIFAR100 on colab - run it tomorrow if possible so we have results by tomorrow}

As elucidated upon in Sec. \ref{sec: intuition} and shown in Sec. \ref{sec: n-step}, the reason behind SAM's robustness is that $\epsilon_1$ is always on the boundary of the neighborhood. By focusing on a boundary point, SAM penalizes high loss only at the boundary, which is characteristic of sharp minima. This selective penalization makes the first-order SAM formulation more robust to isolated spikes in training loss or training noise, effectively ignoring outliers within the $\rho$-radius neighborhood. 

To substantiate this claim from an empirical perspective, we hypothesize that a modification of the SAM algorithm where epsilon is calculated using a unit vector in a random direction instead of the vector along the direction of the loss gradient and scaled by $\rho$ should also generalize better than SGD. We call this modified algorithm Rand-SAM, and we indeed observe that the accuracy of Rand-SAM is better than SGD. Furthermore, what is interesting is that, despite the perturbation being in a random direction, we can see that our Rand-SAM performs comparably to SAM, further highlighting the 'magical' effects of the first-order formulation.

Below in Table. \ref{tab:rand-sam}, we use the above Rand-SAM formulation ($\rho = 0.05$) with WideResNet-28-10 on the CIFAR-10 and CIFAR-100 datasets \cite{alex2009learning} for 200 epochs and compare its performance with SGD and SAM \footnote{SGD and SAM results are from \cite{sam}}.

% \siddarth{Need to cross-check the augmentation value for SGD as proposed in SAM compared to what we do}\\
% \siddarth{Need to recheck the section overall - but specifically I need to recheck the values once more in the table}

% \siddarth{Need to add Cifar-100 Rand-SAM Mean and SD values - below mentioned value is for a single run}

\begin{table}[ht]
    \centering
    \caption{Rand-Sam experiment on CIFAR-10 and CIFAR-100 compared with SGD}
    \label{tab:rand-sam}
    \begin{tabular}{cccc}
    \hline
        \textbf{Model} & \textbf{Dataset} &  \textbf{Optimizer} & \textbf{Accuracy}\\
    \hline
        WideResNet-28-10 &  CIFAR-10 &  SGD &  $96.5_{\pm0.1}$\\
        WideResNet-28-10 &  CIFAR-10 &  \textbf{Rand-SAM} &  $96.9_{\pm<0.1}$\\
        WideResNet-28-10 &  CIFAR-10 &  SAM &  $97.3_{\pm0.1}$\\ \midrule
        WideResNet-28-10 &  CIFAR-100 &  SGD &  $81.2_{\pm0.2}$\\
        WideResNet-28-10 &  CIFAR-100 &  \textbf{Rand-SAM} &  $82.3_{\pm0.1}$\\
        WideResNet-28-10 &  CIFAR-100 &  SAM &  $83.5_{\pm0.2}$\\
    \hline
    \end{tabular}
\end{table}

\section{Limitations and Future Work}

Due to computational resource restrictions, we were unable to run multiple runs of CIFAR-100 of the $N$-step Gradient Ascent experiments in Sec. \ref{sec: n-step}. While we have a few iterations run, we decided not to include the data points since the standard deviation and mean values we report must be statistically accurate, and given the low number of runs, that may not be the case. 

Furthermore, while we have initial results to substantiate our claims, we have run it only on one model for two datasets, the immediate next step for us would be to run multiple iterations of the experiments on several models and a few more datasets. We would also experiment with different augmentation techniques within the dataset. Also, while we substantiate our hypothesis using Rand-SAM, we have not compared the sharpness between SGD and Rand-SAM, which is needed to show the flatter minima. We have only been able to show better accuracy and better generalization ability.

Moreover, on the experimental front of the next steps, it would be helpful to experiment with varying values of $\rho$ on Rand-SAM.
Theoretically, while we have provided mathematical intuition for the 'magic' behind first-order SAM, the next step would be to engage in more rigorous analysis to conclusively prove our intuitions to be true.

\section{Conclusion}

Through this paper, we analyzed the underlying mechanisms behind the generalization success of SAM, explicitly addressing the observed phenomenon of the generalization success of the 1st-order formulation of SAM. We pinpointed the critical role the 1st-order Taylor expansion approximation plays in achieving superior generalization by focusing on boundary points of the neighborhood. Contrary to initial expectations, improving the accuracy of these approximations does not necessarily lead to better generalization, indicating that the effectiveness of SAM lies not in precise sharpness minimization but rather in how it selectively penalizes sharp regions of the loss landscape.

We empirically validate our intuitions through experiments. Utilizing $N$-step gradient ascent, we showed increasing $N$ leads to sharper solutions when compared through a lens of a standardized notion of sharpness. Furthermore, we show that our proposed modification, Rand-SAM, which utilizes a random direction instead of the loss gradient, also demonstrated better generalization than SGD, further supporting our hypothesis that SAM’s generalization performance benefits from boundary-point-based penalization.

While our findings reveal insights into why SAM works, there is still more to explore regarding its mechanisms. Future work could include a more rigorous mathematical analysis to validate our intuitions, testing on additional models and datasets, and investigating the effect of different values of $\rho$ on Rand-SAM. Understanding these aspects will provide a more complete picture of why first-order SAM is effective and will further demystify the mechanisms behind the generalization success of SAM.

% \bibliographystyle{plainnat}
% \bibliography{paper}

\end{document}